\documentclass[cameraready]{Interspeech}
\title{Perceptual compensation for tonal context in self-supervised speech models}
\author[affiliation={1}, orcid=0000-0002-0502-5245]{James}{Kirby}
\author[affiliation={1}]{Ioana}{Krehan}
\author[affiliation={1}]{Michele}{Gubian}
\address{
    $^1$ Institute for Phonetics and Speech Processing, LMU Munich, Germany 
}

\email{jkirby@phonetik.uni-muenchen.de,Ioana.Krehan@campus.lmu.de,m.gubian@phonetik.uni-muenchen.de}


\usepackage{CJKutf8}
\usepackage{comment}
\usepackage[english]{babel}
\usepackage{ipa}

\begin{document}

\maketitle

\begin{abstract}
    This study examines the extent to which the wav2vec2.0 architecture exhibits evidence of compensation for phonological context. We conducted a pseudo-replication of a perceptional compensation experiment on Mandarin Chinese tones, and compared the embedding similarities and probing classifier outputs between a purely self-supervised pre-trained model and a model fine-tuned for Mandarin ASR. No evidence of compensation was found in the embedding similarities of the purely pre-trained model. Probing classifiers showed some evidence of compensation in addition to the expected layer-wise improvements in categorization, but failed to replicate human performance on isolated test syllables. Our findings contrast with previous reports of sensitivity to phonological structure emerging through pre-training alone, and suggest that supervised objectives may be necessary to encourage the abstraction of at least some types of phonological regularities. 
\end{abstract}

\begin{CJK*}{UTF8}{gbsn}

\section{Introduction}

This paper aims to contribute to our understanding of the role of phonological context in speech perception through computational modeling. We do so by exploring how one neural speech architecture, wav2vec2.0 \cite{baevski2020wav2vec2}, represents lexical tones in different phonological contexts. 
By treating wav2vec2.0 as a computational model of perceptual inference and comparing its behavior to that of human listeners, we gain insight into the extent to which these types of models encode phonological structure and what kind of information they use to do so.

\textit{Perceptual compensation} (PC) refers to the phenomenon of context-dependent variation in phonetic category perception, whereby listeners assign acoustically identical 
speech signals to different categories depending on the phonetic or phonological context \cite{sonderegger2010rational}. 
For example,  a following rounded vowel biases fricative perception towards /s/ and away from /\textipa{S}/, despite the fact that the anticipatory lip rounding makes the fricative acoustically more similar to /\textipa{S}/ \cite{mann1980vocalic,mitterer2006causes}. Such context effects have been documented for many speech sounds, in infants \cite{fowler1990young}, and even in nonhuman animals \cite{lotto1997perceptual}, leading to a debate regarding the extent to which the effect can be explained by low-level phonetic mechanisms alone \cite{diehl2004speech,samuel2011speech}. 

Modern neural speech recognition systems are often based on models which have been pre-trained using self-supervised learning (SSL).  
In this framework, models are first pre-trained using a pretext task based on audio only, and then fine-tuned for specific tasks using supervised learning \cite{mohamed2022self}.
The success of the SSL paradigm has spurred much interest in understanding the kinds of linguistic information implicitly encoded in the representations learned by this class of models  \cite{wells2022phonetic,pasad2024what,yang2023what,shen2024encoding, de2024layer}, as well as how model behavior compares with that of humans \cite{scharenborg2018visualizing,millet2022selfsupervised,pouw2024perception,de2024human}.
For example, \cite{scharenborg2018visualizing} show how a supervised ASR model adapts phoneme category boundaries in response to ambiguous training examples in a manner similar to that of humans, while \cite{pouw2024perception} show that a wav2vec2.0 model fine-tuned for English ASR compensates for assimilation more often in phonotactically viable contexts than in non-viable ones. These findings can be interpreted as evidence that neural speech models have acquired knowledge of how speech sounds are realized in different phonological contexts, at least when they have been fine-tuned for tasks involving explicit linguistic categories.

Arguably more surprising are claims that purely pre-trained models also display such behavior. 
Like \cite{pouw2024perception}, \cite{de2024human} investigated how wav2vec2.0 encodes ambiguous speech sounds in phonotactically viable and non-viable contexts. They observed an effect of context in both pre-trained and fine-tuned models, leading them to suggest that the model can implicitly learn English phonotactic structure even without a symbolic training objective. Such findings are striking, because they suggest that the context encoder networks of SSL models may be implicitly capturing information about phonological context without being given any explicit information about phonological categories. This would be consistent with accounts of PC that are based on purely phonetic knowledge \cite{gow2003feature,gow2004crosslinguistic}.

In this paper, we contribute to this line of research by testing the extent to which a neural speech architecture compensates for context in the domain of lexical tone, through a pseudo-replication of a psycholinguistic study of PC in Mandarin Chinese \cite{zhang2022influence}. We hypothesize that the contextualized speech embeddings learned with self-supervision may be especially adept at compensating for suprasegmental contexts such as tone. 
We focus here on the wav2vec2.0 architecture, as it has been widely used in previous related work and because both pre-trained (PT) and fine-tuned (FT) checkpoints exist for Mandarin Chinese. 
We analyze how the model encodes semi-controlled psycholinguistic stimuli through the use of probing classifiers and the study of embedding similarities. 

We find that while the embedding similarities of the FT model show some evidence of PC in later network layers, those of the PT model show no evidence of compensation for tonal context at any layer. Probing classifiers tested using both PT and FT embeddings show some indications of PC, but the classification of isolated syllables encoded without context diverges sharply from that of human listeners. Our results therefore suggest caution before concluding that self-supervised pre-training objectives are sufficient for neural speech models to integrate information about phonological category representations; supervised learning may still be necessary to encourage the integration of at least some types of phonological context.

\section{Background}

\begin{figure}[!t]
         \includegraphics[width=\linewidth, trim=10 0 10 10, clip]{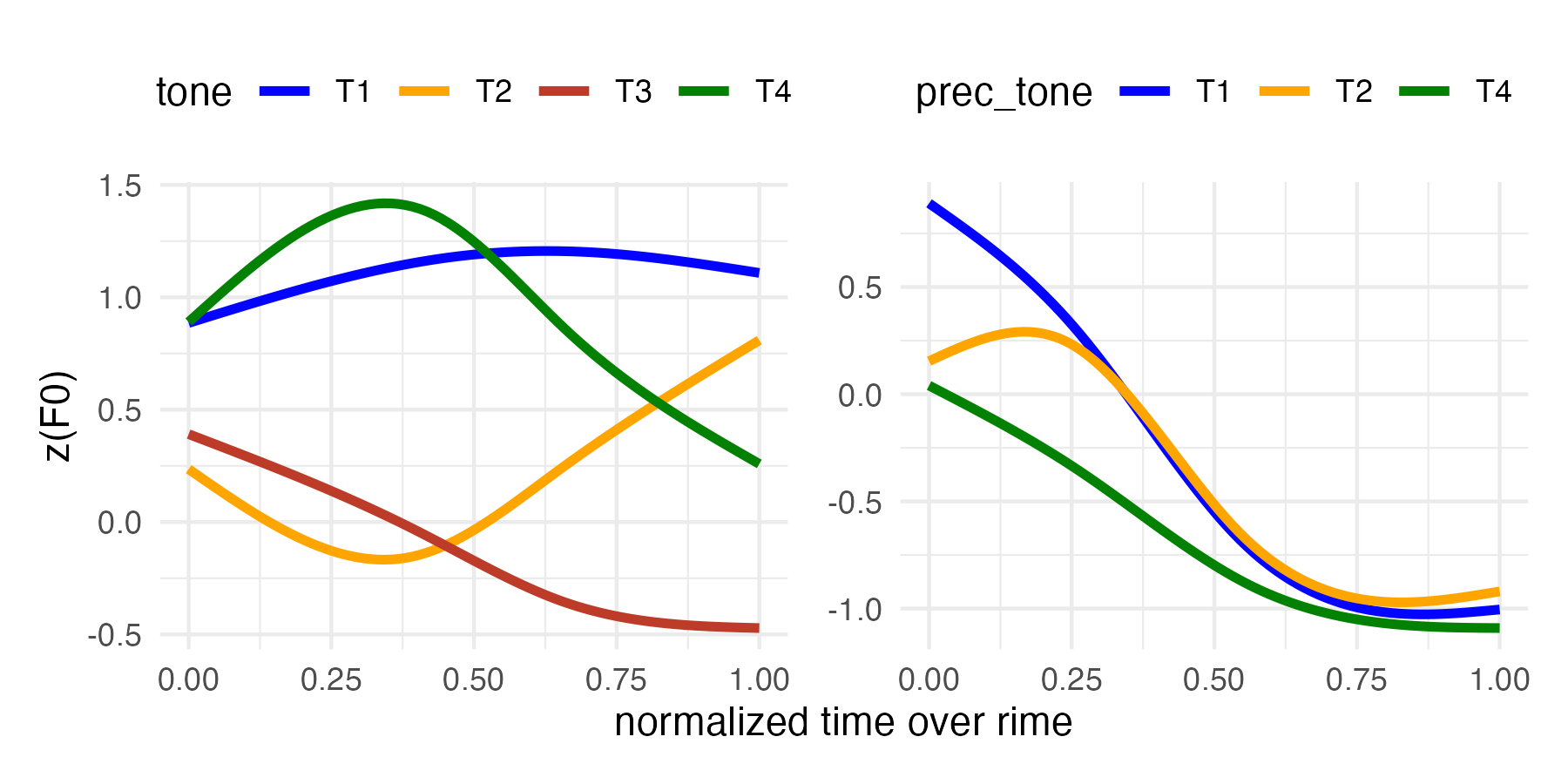}
         \caption{Left: $F0$ contours of the four lexical tones of Mandarin Chinese on sonorant syllables extracted from the AISHELL-3 corpus \cite{shi2021aishell}. Right: realization of Tone 3 in the context of preceding tones.}
         \label{fig:mandarin}
\end{figure}

Most lexical syllables in Mandarin Chinese bear one of four tones: high-level Tone 1 ([ba1] {八} `eight'), high-rising Tone 2 ([ba2] {拔} `to pull'), low-falling(-rising) Tone 3 ([ba3] {把} `to take') or high-falling Tone 4 ([ba4] {爸} `father'). 
Like segments, the phonetic realization of tones is heavily influenced by carryover coarticulation. 
For example, although Mandarin T3 is typically realized with a low-falling contour in post-pausal position (Fig. \ref{fig:mandarin}, left), when preceded by a tone that ends high, like T1, it is realized with a high-falling contour (Fig. \ref{fig:mandarin}, right) \cite{xu1994production}. 
In other words, when following a T1 or T2, the onset of T3 is raised, increasing its acoustic similarity to canonical T4; similarly, in the context of a preceding T4, the onset of T4 is lowered, increasing its acoustic similarity to canonical T3.

In perception, however, human listeners compensate for these coarticulatory effects 
\cite{zhang2022influence,xu1994production,chen2016context,fox1990context}. \cite{zhang2022influence} demonstrated the effect of preceding tonal context on the perceptual boundary between low-falling T3 and high-falling T4 by having participants classify stimuli drawn from a 14-step T4-T3 continuum using a two-alternative forced choice paradigm. Each trial consisted of a \textit{target} stimulus either in isolation, or preceded by one of the three \textit{context} syllables. Results (Fig.~\ref{fig:Zhang2022results}) showed a clear preference for T3 responses when the preceding tone was high-offset T1 or T2, and T4 when the preceding tone was low-offset T4. 
Importantly, stimuli presented in isolation (labeled `no-ctx' in Fig.~\ref{fig:Zhang2022results}) were intermediate, suggesting that the tonal context serves to bias a general phonological representation. 
\section{Methods}

\subsection{Stimuli}
\label{sec:stimuli}

To investigate whether wav2vec2.0 displays a similar bias in its responses, we generated continua of Mandarin syllables with fixed segmental material but $F0$ contours varying between canonical T3 and T4 endpoints.
Typical for psycholinguistic studies, \cite{zhang2022influence} tested a small number of stimuli on a large number of participants. 
Since model responses to individual stimuli will be deterministic, we introduced variability by resynthesizing a large number of tonal continua with different preceding tone contexts by extracting disyllable sequences from forty speakers from the \emph{test} split of the AISHELL-3 corpus \cite{shi2021aishell}, using the Montreal Forced Aligner \cite{mcauliffe17_interspeech} to determine syllable boundaries.

\begin{figure}[!ht]
         \centering
         \includegraphics[width=0.8\linewidth]{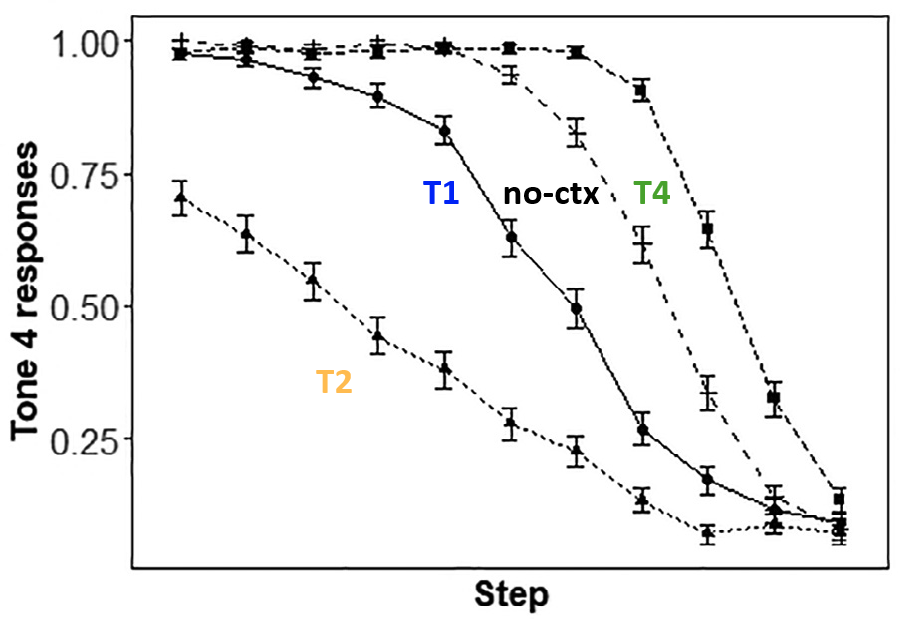}
         \caption{Proportion of T4 responses from human listeners in Experiment 3 of \cite{zhang2022influence}.}
         \label{fig:Zhang2022results}
\end{figure}

As in \cite{zhang2022influence}, \textit{context} syllables
bore one of T1, T2 or T4, while \textit{target} syllables bore
either T3 or T4. 
To exclude creaky samples, we analyzed the $F0$ of potential target audio clips and selected only those where at least 10 $F0$ samples were present and above the 10\% quantile of $F0$ for the corresponding speaker. We then used \textit{Parselmouth} \cite{parselmouth, praat} to apply the duration and $F0$ manipulations described in \cite{zhang2022influence} to each context-target disyllable to obtain a 14-step T4-T3 target continuum preceded by either a context syllable (with T1, T2, or T4) or in isolation (labeled as \textit{no-context}). 
Speaker-dependent 10\% and 90\% quantiles were used to determine $F0$ extrema for the continuum endpoints. This procedure generated around 13,700 continua (ca.~192,000 stimuli). 

\begin{figure*}
         \includegraphics[trim = 0 0 0 20, width=1.0\linewidth]{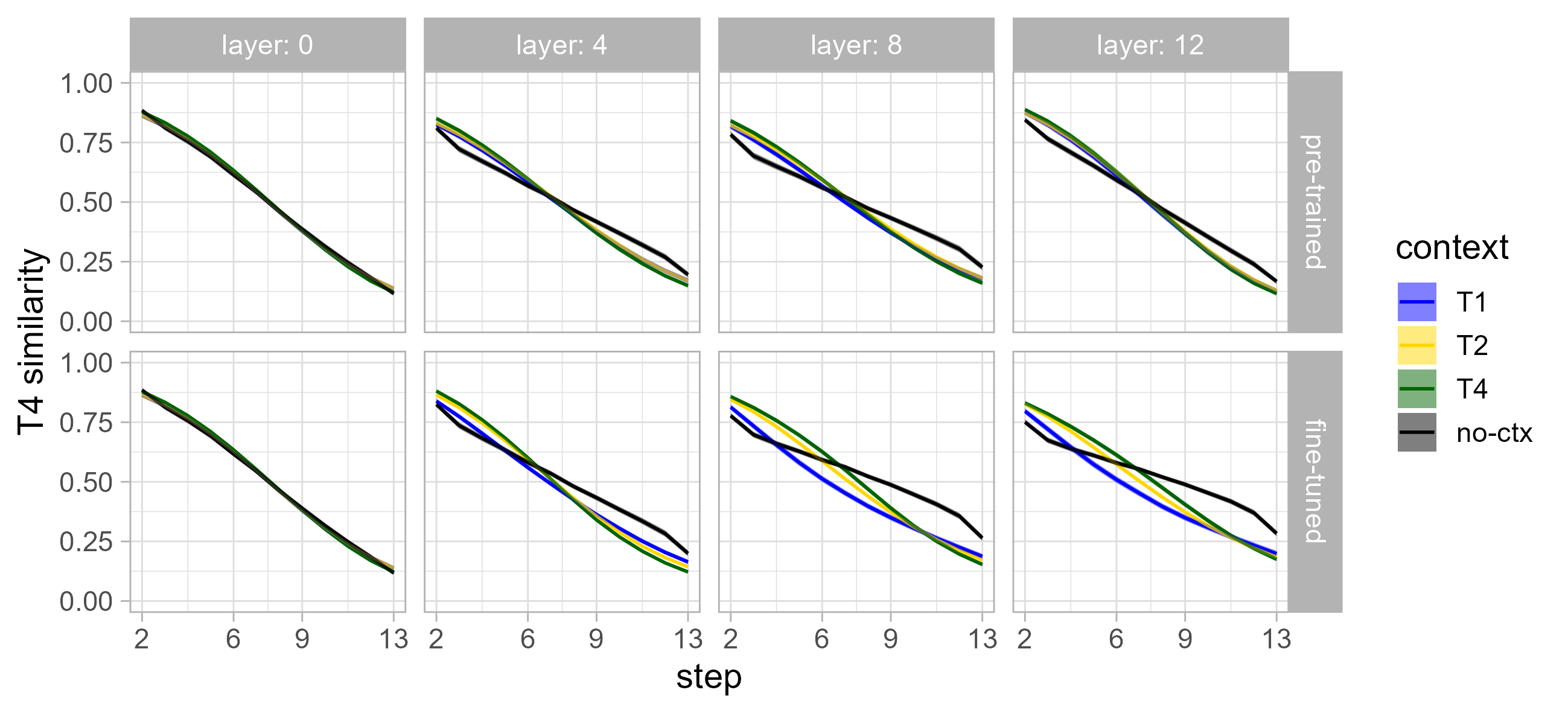}
         \caption{T4 embedding similarities \eqref{equation:sim} as function of T4-T3 continuum step; line thickness indicates 95\% confidence bands.}
         \label{fig:emb_sim}
\end{figure*}

\subsection{Model checkpoints}

All stimuli were then processed by two wav2vec2.0 model checkpoints: one pre-trained on 1,000 hours of untranscribed Mandarin speech only\footnote{\url{https://github.com/kehanlu/mandarin-wav2vec2}}, and one fine-tuned for the task of Mandarin ASR using 178 hours of transcribed speech\footnote{\url{https://huggingface.co/kehanlu/mandarin-wav2vec2-aishell1}}. Both models contain 7-layer CNN feature encoders followed by 12 Transformer layers. We used the Transformers library \cite{wolf2020transformers} to download and process both checkpoints.

\subsection{Analysis methods}

We report results of two methods for analyzing SSL model representations: embedding similarities and probing classifiers.  
We applied both methods to the 512-dimensional output of the convolutional feature extractor (layer 0) as well as to the 768-dimensional outputs of the Transformer (layers 1-12). 

\subsubsection{Embedding similarities}\label{sec:method.emb}

We inspected the internal representations of both models by comparing  the embedding similarities of each stimulus step with the corresponding T3 and T4 endpoints (cf. \cite{de2024human}). 
For a given stimulus $X$, the corresponding manipulated step-1 (T4) and step-14 (T3) endpoints based on the same source syllable were used as reference stimuli. For each stimulus, vectors were derived by averaging layer-wise embeddings over the entire syllable.
Following \cite{de2024human}, we then computed the similarity of $X$ to T4 relative to T3 as:
\begin{equation}
    sim(X, \text{T4}) = 1 - \frac{D_{cos}(X, \text{T4)}}{D_{cos}(X, \text{T4)} + D_{cos}(X, \text{T3)}},
    \label{equation:sim}
\end{equation}
where $D_{cos}(a, b) = 1 - cos(a,b)$. Similarities were modeled as a continuous, [0,1] bounded beta-distributed response variable in a generalized additive mixed model (GAMM) using the \textit{mgcv} R library \cite{wood2011fast}.
Four smooth functions of \textit{step} model the effect of \textit{context} (T1, T2, T4 or \textit{no-ctx}), where \textit{step} ranges between 2 and 13 (extremes are excluded as they coincide with the references 1=T4 and 14=T3). 
Speaker-by-context random smooth terms were added to control for speaker variability. One GAMM was fit independently for each checkpoint and layer.

\subsubsection{Probing classifiers}
\label{sec:probes}

We trained probing classifiers to predict T3/T4 tone labels from the representations extracted from each model checkpoint and layer, following \cite{ma2021probing}.
As for the similarity computation, vectors were obtained by averaging embeddings over target syllables.

Probing classifiers were linear, fully connected neural networks implementing binary logistic regression with cross-entropy loss, using the Adam optimizer with learning rate $10^{-3}$. 
Classifiers were trained on sets of T3 and T4 syllables from forty speakers from the \textit{train} split of the AISHELL-3 corpus. We trained on 100 T3 and 100 T4 syllables for each of 36 speakers (18 female) and validated on 4 speakers (2 female). 
Training was run for five epochs. Validation accuracy saturated between 82\% in the CNN layers and up to 99\% in the higher Transformer layers. Errors were moderately imbalanced in favor of false T3 recognition, the most extreme being layer 12, where 4-6 times more false T3s than false T4s were predicted.

Testing was carried out on the classifier state after the last training epoch.
The test set consisted of the manipulated stimuli described in Sec.~\ref{sec:stimuli}. 
Binary results are modeled as a Bernoulli-distributed response variable in GAMMs with the same predictor structure described in Sec.~\ref{sec:method.emb} (step extremes included).

\begin{figure*}[htbp!]
         \includegraphics[trim = 0 0 0 20, width=1.0\linewidth]{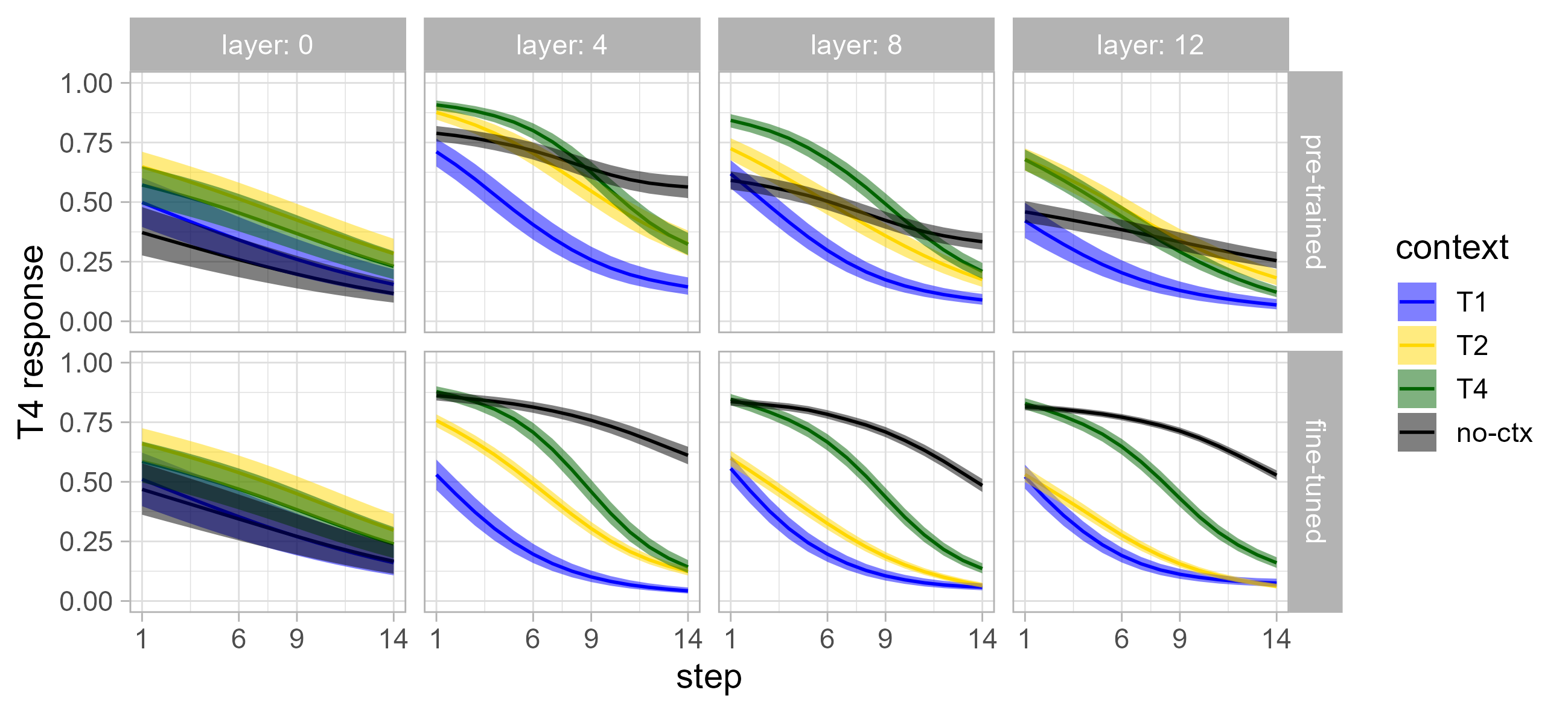}
         \caption{Proportion of T4 responses of probes as function of T4-T3 continuum step; line thickness indicates 95\% confidence bands.}
         \label{fig:probes}
\end{figure*}

\section{Results}

Here we only show the results for layer 0 (CNN output) and layers 4, 8, and 12; results for other layers follow similar trends.

\subsection{Embedding similarities}

Embedding similarities for both PT and FT checkpoints are shown in Fig.~\ref{fig:emb_sim}. There is no evidence of sensitivity to context in the PT embeddings (top row). There appears to be a small effect of the addition of context (compared to the no-context condition), but not of compensation in the sense seen in \cite{zhang2022influence}. 
The embedding similarities from the FT model (bottom row), on the other hand, do show some evidence of context sensitivity: ambiguous stimuli in a T1 context have contextual representations that are more similar to T3 than to T4, indicating that fine-tuning fundamentally alters the model’s internal representations. However, these shifts are quite small compared to the human responses, and again do not seem to be relative to the no-context embeddings. In addition, the contexts T2 and T4 appear to pattern together relative to T1, with T1 showing the greatest shift, which is again a qualitatively different pattern compared to \cite{zhang2022influence} (Fig.~\ref{fig:Zhang2022results}). 

\subsection{Probing classifiers}

Fig.~\ref{fig:probes} shows the proportion of T4 responses at each continuum step for PT and FT models at four layers. 
A mild effect of context is visible at the final CNN layers, but already in the fourth Transformer layers we see an enhanced sensitivity to the continuum endpoints together with sensitivity to context. This is strongest in layer 8 of the FT model (bottom row, third panel) where the response patterns most qualitatively resemble those of human participants (Fig.~\ref{fig:Zhang2022results}). 

However, in the no-context condition, the expected sigmoidal response curve fails to emerge in either model at any layer. This is most visible in the probes tested on FT embeddings, where we observe a strong bias for T4 responses regardless of the $F0$ contour of the stimulus.

\section{Discussion}

This study explored the extent to which two wav2vec2.0 models, one pre-trained and one fine-tuned, exhibited evidence of compensation for phonological context. Purely self-supervised representations showed no evidence of the baseline-centered boundary shifts that characterized perceptual compensation for tonal coarticulation in \cite{zhang2022influence}. While some evidence of compensation was seen in the later layers of the fine-tuned model, fine-tuning induced only weak, qualitatively non-human-like shifts that are not clearly anchored to the no-context baseline. 

The lack of evidence for PC in the purely self-supervised model was inconsistent with our hypothesis that Transformer-based embeddings would have an advantage at inducing contextually specific tonal representations, as well as with previous reports of phonological encoding emerging in purely self-supervised models \cite{ shen2024encoding,de2024layer,de2024human,kloots2025what,bentum2025word}. 
Indeed, our most robust finding is the complete lack of context sensitivity in purely pretrained embeddings (Fig.~\ref{fig:emb_sim}, top). 

To what should we attribute these differences? One possibility is that the acoustic realizations of the segmental [r/l] contrast studied by both \cite{scharenborg2018visualizing} and \cite{de2024human} are overall less variable across speakers and contexts than lexical tones, which are realized primarily through modulation of $F0$, a trajectory subject to perturbation via baseline shifts, range scaling, truncation, intonational trends, and local segmental perturbations. 
In other words, the set of acoustic dimensions that strongly affect the acoustic signal for tone may be much larger than for something like the [r/l] contrast, because $F0$ is jointly determined by a large number of extraneous factors. 
From the model's perspective, the influence of the preceding tonal context is competing with many other plausible explanations to account for the observed $F0$ trajectory.

Both the embeddings of the FT model, as well as the linear probes fit to the embeddings of both models, showed some evidence of PC. We attribute this to the fact that both the fine-tuning objective and the probing classifiers involving supervised training, where the model or classifier is explicitly told of the existence of lexical tone categories. 
However, while the probing results showed the expected increase in accessibility of tone categories across layers, the responses to isolated syllables still diverged sharply from that of human listeners. This finding highlights a potentially important difference between human and machine processing of phonological context: the representations learned by even fine-tuned models seem to rely on broader contextual conditioning, rather than supporting a context-invariant tonal category 
biased by local context.

However, we must acknowledge some limitations of our probing results. 
Our probes were trained on instances of T3 and T4 extracted from embeddings made from whole utterances, encoding rich contextual information, whereas they were tested on embeddings made from isolated syllables. Specifically for the task of single-syllable prediction, the utterance-level context encoded by the FT model may actually be detrimental for the specific task of syllable-level tone identification \cite{robertson2023bigger,meng2025effective}.
While we were careful to insure our probes did not encode a frequency bias (Sec.~\ref{sec:probes}), we nonetheless observed a strong bias for T4 when classifying isolated syllables using the FT model. We have no explanation for this at present, but speculate that this may reflect the fact that T4 is the most frequent tone in Mandarin speech \cite{wu2020mandarin}. Moreover, for many speakers, a canonical pre-pausal T3 will be realized with a final rise, which was never present in our stimuli (or those of \cite{zhang2022influence}).

On the other hand, our probes were more accurate than the embedding similarities when classifying our disyllabic test stimuli, even for the PT model.
Although it is tempting to interpret high probing accuracy as indicating the model `encodes' tonal context,
it is more accurate to say that it indicates that the model has learned a representation in which the probed categories are \textit{accessible}.
If the categories in question correlate with other properties in the training data, even a simple linear model may be able to find a fit to subtle correlations that do not directly reflect the property of interest \cite{belinkov2022probing}.
In future work, we intend to supplement our analysis measures with causal interchange interventions \cite{pouw2024perception} and task-based evaluations \cite{de2024human}, as well as considering the effects of nonspeech analogs as in \cite{zhang2022influence}.

In summary, our results highlight a dissociation between contextualization and perceptual compensation. Purely self-supervised wav2vec2.0 representations showed no evidence of compensation for tonal coarticulation, and even supervised probes fit to fine-tuned representations diverged sharply from responses of human listeners on isolated syllables, suggesting that the learned representations rely on broader contextual conditioning rather than biasing a context-invariant baseline. Taken together, these results constrain accounts of perceptual compensation by indicating that, at least for tones,  human-like PC patterns are unlikely to fall out of unsupervised contextual prediction alone, but instead require additional learning mechanisms that encourage the emergence of stable phonological category representations. 

\section{Generative AI Use Disclosure}
No generative AI or AI-assisted technologies were used in the research process or the preparation of this manuscript.

\bibliographystyle{IEEEtran}
\bibliography{mybib}

\end{CJK*}

\end{document}